\def\year{2022}\relax
\documentclass[letterpaper]{article} 
\usepackage{aaai22}  
\usepackage{times}  
\usepackage{helvet}  
\usepackage{courier}  
\usepackage[hyphens]{url}  
\usepackage{graphicx} 
\usepackage{multirow}
\usepackage{booktabs}
\usepackage{subfigure}
\usepackage{natbib}  
\usepackage{caption} 
\DeclareCaptionStyle{ruled}{labelfont=normalfont,labelsep=colon,strut=off} 
\usepackage{algorithm}
\usepackage{algorithmic}
\usepackage{amsmath}
\usepackage{threeparttable}
\usepackage{amssymb}
\usepackage{newfloat}
\usepackage{listings}
\floatstyle{ruled}
\newfloat{listing}{tb}{lst}{}
\floatname{listing}{Listing}
\title{Spiking CapsNet: A Spiking Neural Network With A Biologically Plausible Routing Rule Between Capsules }
\author{
    Dongcheng Zhao\textsuperscript{\rm 1,3} \equalcontrib,
    Yang Li\textsuperscript{\rm 1,2,3}\equalcontrib,    
    Yi Zeng\textsuperscript{\rm 1,2,3,4,5}
\equalcontrib
    \thanks{Corresponding Author.},
    Jihang Wang\textsuperscript{\rm 1,2,3},
    Qian Zhang\textsuperscript{\rm 1,2,3}
}
\affiliations{
    \textsuperscript{\rm 1}Institute of Automation, Chinese Academy of Sciences (CASIA), Beijing, China\\
    \textsuperscript{\rm 2}School of Artificial Intelligence, University of Chinese Academy of Sciences, Beijing, China\\
    \textsuperscript{\rm 3}Research Center for Brain-Inspired Intelligence, CASIA, Beijing, China\\
    \textsuperscript{\rm 4}National Laboratory of Pattern Recognition, CASIA, Beijing, China\\
    \textsuperscript{\rm 5}Center for Excellence in Brain Science and Intelligence Technology, \\Chinese Academy of Sciences, Shanghai, China \\

    \{zhaodongcheng2016, liyang2019, yi.zeng\}@ia.ac.cn
}

\usepackage{bibentry}

\begin{document}

\maketitle

\begin{abstract}
Spiking neural network (SNN) has attracted much attention due to their powerful spatio-temporal information representation ability. Capsule Neural Network (CapsNet) does well in assembling and coupling features at different levels. Here, we propose Spiking CapsNet by introducing the capsules into the modelling of spiking neural networks. In addition, we propose a more biologically plausible Spike Timing Dependent Plasticity routing mechanism. By fully considering the spatio-temporal relationship between the low-level spiking capsules and the high-level spiking capsules, the coupling ability between them is further improved. We have verified experiments on the MNIST and FashionMNIST datasets. Compared with other excellent SNN models, our algorithm still achieves high performance. Our Spiking CapsNet fully combines the strengthens of SNN and CapsNet, and shows strong robustness to noise and affine transformation. By adding different Salt-Pepper and Gaussian noise to the test dataset, the experimental results demonstrate that our Spiking CapsNet shows a more robust performance when there is more noise, while the artificial neural network can not correctly clarify. As well, our Spiking CapsNet shows strong generalization to affine transformation on the AffNIST dataset. 
\end{abstract}
\section{Introduction}
Convolutional Neural Networks (CNNs) have obtained tremendous success in various domains, such as object classification~\citep{he2016deep}, visual tracking~\citep{danelljan2015convolutional}, object segmentation~\citep{chen2017deeplab}, and so on due to the powerful feature representation ability. However, there still exist several limitations associated with CNNs. First, high-level features are obtained by low-level weighting features while ignoring the spatial relationship between them. A face is just a simple combination of the features of eyes, nose, and mouth. Even if the positions of these features change relatively, the classifier will still consider it to be a face. Secondly, the commonly used pooling operation is destructive to the spatial relationship. To tackle the problems, CapsNet~\citep{sabour2017dynamic,hinton2018matrix} proposes the concept of capsules, which uses the vector or matrix instead of values to encode the information. It uses a group of neurons to denote more properties. Each neuron provides a scalar output that represents the attributes of the corresponding feature, such as position, color, and texture. The length
of the vector represents the probability of these properties. In addition to the capsule concept, the routing scheme is used to ensure the output of the lower-capsule to the closely related higher-capsule.  

Although CapsNet mimics the multi-layer visual system with a parse tree-like structure, it is still far from the human brain's information processing mechanism. Considered as the third-generation artificial neural network, the spiking neural networks (SNNs)~\citep{roy2019towards,kim2019spiking,thiele2019spiking,zhang2019tdsnn},
use the discrete spikes to transfer information, which is more biologically plausible and more energy efficient. The SNNs present strong sparsity and neurons that do not emit spikes do not update the network weights. In recent years, SNNs have significantly facilitated the development of event-based neuromorphic hardware platforms\cite{pei2019towards}, showing desired low latency and high efficiency. On the other hand, due to the spike encoding mechanism, the SNNs are very robust to the input disturbance and have shown excellent performance in terms of anti-noise\cite{cheng2020lisnn,chowdhury2020towards,zhang2019fast,li2020robustness,uysal2007spike}. Many traditional SNN structures are inspired by brain area connections to imitate certain brain area functions. Still, it is hard to expand to a deeper network structure due to its complex network connections~\citep{zhao2020neural,zhao2018brain,fang2021brain}. To carry out more complex tasks, the feedforward fully connected structure is introduced~\citep{zhang2018plasticity,sun2021quantum} to transmit information in the form of a fully connected way. However, this connection method will cause excessive parameters and overfitting, which is difficult to extend to deeper neural networks and more complex tasks. To tackle the problems, GLSNN~\citep{zhao2020glsnn} introduces the feedback connections to transfer the global error to the hidden layers to get the corresponding target. Combining with the local synaptic plasticity learning rules, the performance and stability are further improved. Convolutional neural networks have attracted wide attention due to their superior feature extraction capabilities. Recent works~\citep{wu2018spatio, wu2019direct, jin2018hybrid, zhang2020temporal} introduce the convolutional structures into the modelling of SNNs. The parameter sharing mechanism of convolutions dramatically reduces the number of parameters and can further deepen the structure of the SNNs and improve the performance of complex tasks. LISNN~\citep{cheng2020lisnn} further enhances the performance of the SNNs and their robustness to noise by introducing the lateral connections. BackEISNN~\citep{zhao2021backeisnn} has demonstrated superior performance on multiple datasets by introducing self-feedback connections and the excitatory-inhibitory neurons. However, these structures all use pooling operations to ensure translation invariance, leading to the spatial information loss. To tackle the problems mentioned above, we introduce the capsule structure in the spiking neural network, so as to take full advantage of the  spatial and temporal characteristics.

In the capsule neural network domains, much work is integrated into the modelling of routing mechanisms, which can be roughly divided into two categories, the supervised and the unsupervised. 

For the unsupervised ones, the dynamic routing~\citep{sabour2017dynamic} uses the inner product to denote the agreement. Furthermore, the vector capsule is replaced with the matrix, and the modified EM-algorithm is used to model the agreement between capsules~\citep{hinton2018matrix}. The group equivariant capsule networks~\citep{lenssen2018group} present a generic routing by agreement algorithm defined on elements of a group. \citep{choi2019attention} proposes a fast forward pass routing with attention modules to keep spatial attention. Riberio et al.~\citep{ribeiro2020capsule} propose a routing algorithm derived from Variational Bayes to fit a gaussian mixture model. In~\citep{tsai2020capsules}'s work, the routing procedure resembles an inverted attention algorithm. ~\citep{zhang2021fast} generalizes the existing routing methods within the framework of weighted kernel density estimation.  Efficient-CapsNet~\citep{mazzia2021efficient} replaces the dynamic routing with a novel non-iterative, highly parallelizable self-attention routing.

For the supervised ones, Wang et al.~\citep{wang2018optimization} formulate the routing strategy as an optimization problem that minimizes the distance between the current coupling distribution and its last states. \citep{li2018neural} approximates the routing process with a master and an aide branch to communicate in a fast, supervised, and one-time pass fashion. G-CapsNet~\citep{chen2018generalized} embeds the routing procedure into the optimization procedure and makes the coefficients trainable. Self-Routing~\citep{hahn2019self} routes each capsule independently by its subordinate routing network. STAR-CAPS~\citep{ahmed2019star} designs a straight-through attentive routing by utilizing attention modules augmented by differentiable binary routers. GF-CapsNet~\citep{ding2020group} adopts a supervised group-routing to equally spilled capsules into groups to reduce routing parameters. Both the supervised and the unsupervised ones only consider the spatial relationship between capsules while ignoring the temporal relationship. This paper adopts the biologically plausible Spike Timing Dependent Plasticity (STDP)~\citep{bi1998synaptic,diehl2015unsupervised} for dynamic routing between the spiking capsules, which fully considers the causal relationship between capsules and significantly improves routing efficiency. 

This paper proposes the Spiking CapsNet, which can combine the characteristics of the capsule neural network and the spiking neural network well. And our contributions can be summarized below. 
\begin{itemize}
\item  To our best knowledge, this is the first work to introduce the capsule structures into the modelling of SNNs, which can combine their rich spatio-temporal information processing capabilities.
\item We introduce a more biological routing algorithm between the spiking capsules to optimize the relationship between capsules in both spatial and temporal domains. The routing algorithm fully considers the spatial relationship between the part and the whole and the causality of the spike sequences in temporal domains. 
\item The experimental results show that our proposed Spiking CapsNet does well on the MNIST and FashionMNIST datasets compared with the current best SNN. It can achieve the best noise robustness under different Salt-Pepper noise intensities and different variance of Gaussian noise. As well as, it shows excellent generalization and invariance to affine-transformations. 
\end{itemize}

\section{Methods}
This section will introduce the learning and inference process of our proposed Spiking CapsNet in detail. Firstly, we describe the spiking neuron model, followed by the description of the capsule operation process, which is very different from the traditional value-based operation. Then, we will give a detailed description of our STDP routing mechanism. Finally, the introduction of the whole Spiking CapsNet will be given.

\subsection{Spiking Neuron Model}
\begin{figure}[htbp]
	\centering
	\includegraphics[scale=0.65]{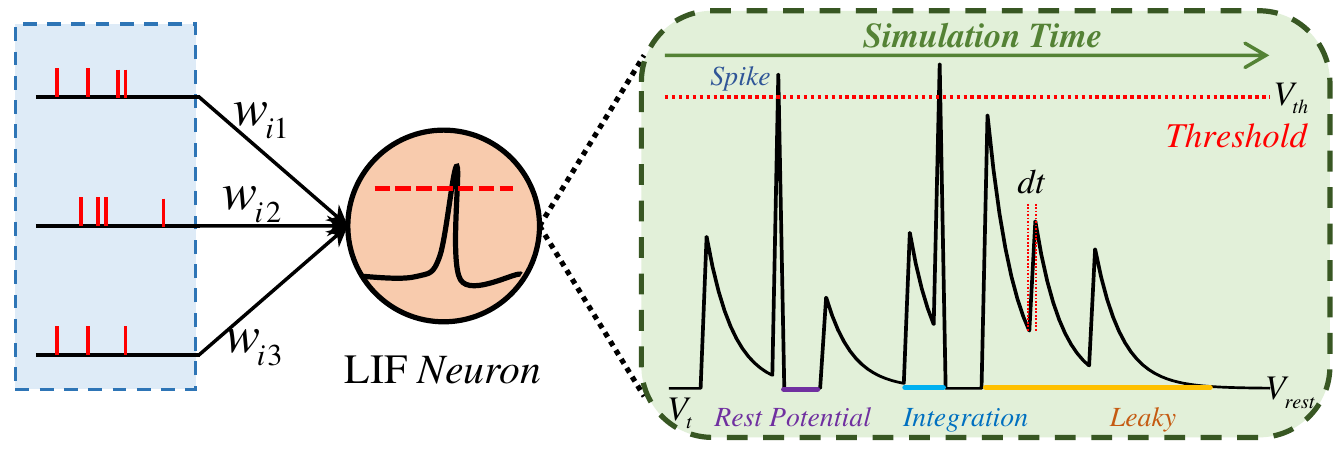}
	\caption{Illustration of LIF Neuron}
	\label{LIF}
\end{figure}
The basic unit in SNNs is the spiking neuron, which is  abstracted from the biological neuron.  Leaky-Integrate-and-Fire (LIF) neuron model is the most commonly used by the deep spiking neural network due to its superficial calculation characteristics, which can be seen in Fig.~\ref{LIF}. The LIF neuron continuously receives spikes $\delta_j$ from the pre-synaptic neuron and dynamically changes its membrane potential $V_i(t)$. When it exceeds the threshold $V_{th}$, the neuron spikes. In order to reduce the information loss and ensure the convergence of the network, we use the soft reset method used in~\citep{rueckauer2017conversion}, when the membrane potential exceeds the threshold, the membrane potential will subtract the threshold, and the details are shown in Eq.~\ref{lif3}.
\begin{equation}
	\left\{
		\begin{aligned}
			C\frac{dV_i(t)}{dt} &= -\frac{1}{R}(V_i(t)-V_{rest})+\sum\limits _{j=1}^{N}w_{ij}\delta_j(t)  \\
			V_i(t) &= V_i(t)- V_{th} \quad\delta_i(t)=1 \quad if \quad V_i(t)\geq V_{th}
		\end{aligned}
	\right.
	\label{lif3}
\end{equation}
where $C$ and $R$ are the membrane capacitance and membrane resistance respectively, $w_{ij}$ is the synaptic weight, $N$ denotes the number of pre-synaptic neurons, $dt$ is the minimal simulation time step for $V_i(t)$, which is usually 1 ms.

In order to facilitate the calculation and modelling, we convert the Eq.~\ref{lif3} into a discrete form as shown in  Eq.~\ref{memupdate}, and $\lambda$ denotes the decay factor of the membrane potential.
\begin{align}
	\label{memupdate}
	V[t] = \lambda V[t-1]+\sum\limits _{j=1}^{N}w_{ij}\delta_j[t]
\end{align}
\subsection{Capsule Neural Networks}
\begin{figure}[htbp]
	\centering
	\includegraphics[scale=0.75]{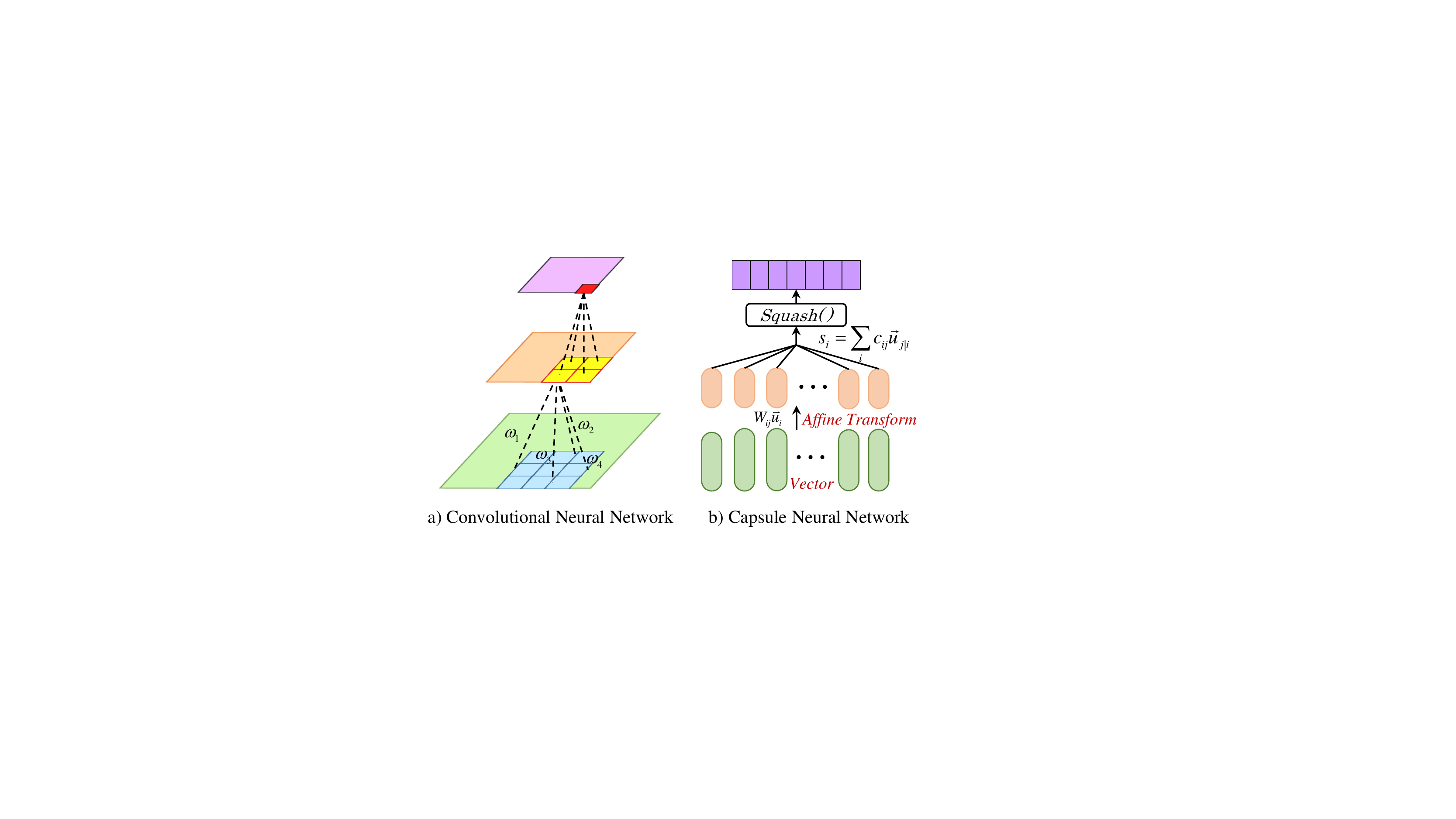}
	\caption{Convolutional Neural Networks vs Capsule Neural Networks.}
	\label{caps}
\end{figure}

There are two main differences between the capsule neural network and the convolutional neural network as shown in Fig.~\ref{caps}. First, the capsule neural network uses a vector to represent the internal properties of an entity. Second, the capsule neural network uses the routing mechanism to replace the pooling layer in the convolution neural network. The low-level capsule output is used as the input of the corresponding high-level capsule to represent the relationship between the part and the whole to achieve the translation equivariance. 

In general, the architectures of capsule networks consist of a traditional convolutional layer to extract features as the input of the initial capsule layer, a PrimaryCaps layer to convert the scalar representations to vector or matrix representations, and the DigitCaps layer to output the final predicted probability of each category.
 \subsection{STDP Routing}
In this subsection, we will describe the STDP routing mechanism used in this paper. First, we review the dynamic routing in the traditional CapsNet. 

\begin{figure}[h]
	\centering
	\includegraphics[scale=0.45]{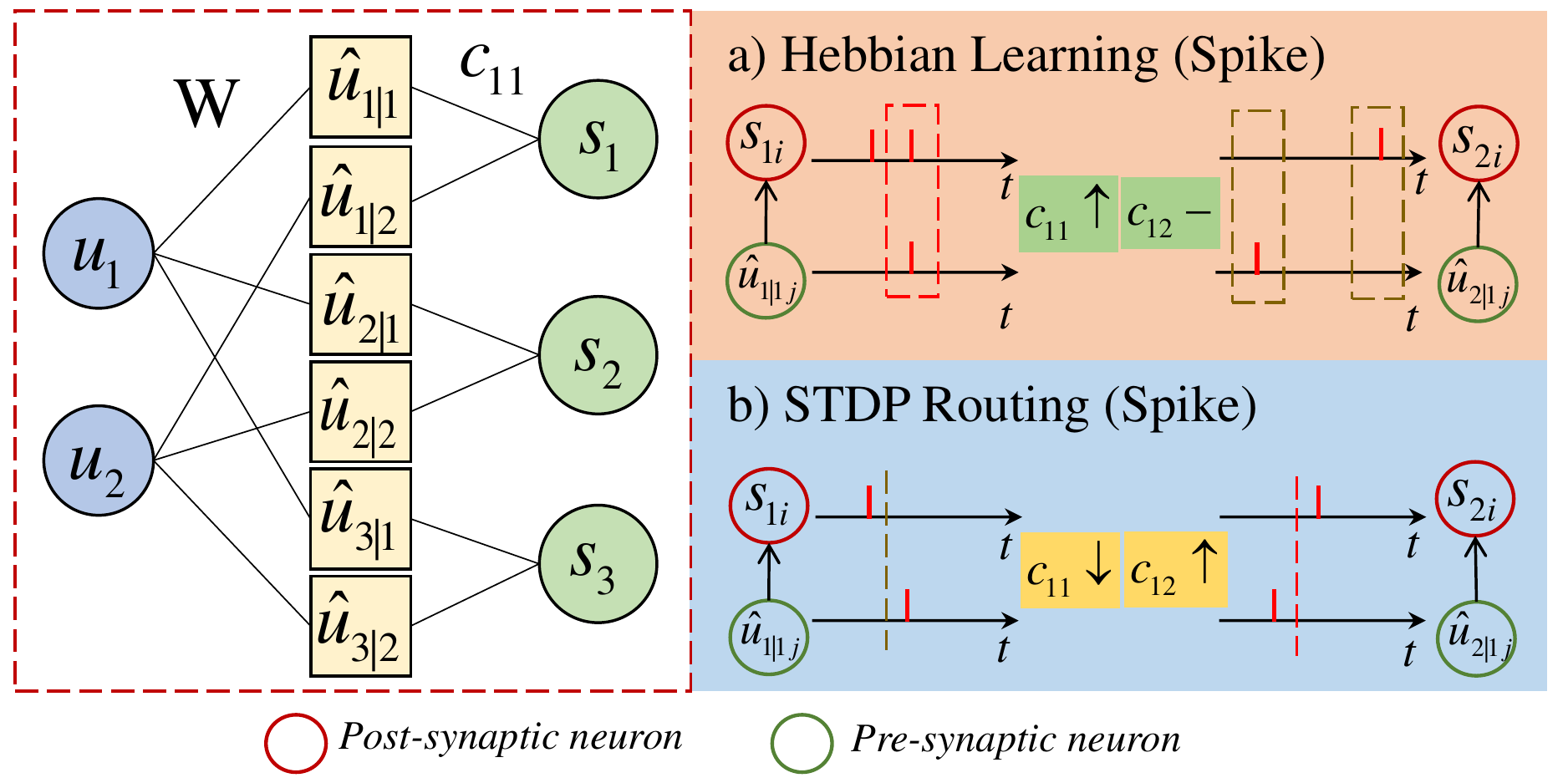}
	\caption{The STDP Routing}
	\label{routing}
\end{figure}

As shown in the Fig.~\ref{routing}, suppose $u_i$, $v_j$ represent the $i_{th}$ low-level and $j_{th}$ high-level capsules respectively. First, through a transformation matrix $w_{ij}$, each low-level capsule will have a prediction vector $\hat{u}_{j|i}$ for the high-level capsule. Then according to the routing coefficient $c_{ij}$, a weighted summation of the predictions is performed to obtain $s_j$. Finally, using a non-linear squash function to shrunk to a length in (0,1), we can get the high-level capsules $v_j$. The details are shown in Eq.~\ref{caps1}.
\begin{equation}
	\left\{
		\begin{aligned}
		&\hat{u}_{j|i} = w_{ij} u_i	  \\
		&s_j = \sum_i c_{ij}\hat{u}_{j|i} \\
		&v_j = \frac{||s_j||^2}{1+||s_j||^2}\frac{s_j}{||s_j||}\\
		\end{aligned}
	\right.
	\label{caps1}
\end{equation}

The routing logits $b_{ij}$ are initialized to 0 and are updated by an unsupervised approach iteratively, the final coefficients $c_{ij}$ are obtained  by applying the softmax operation as shown in Eq.~\ref{caps2}. $r$ is the iteration number. 
\begin{equation}
	\left\{
		\begin{aligned}
		&b_{ij}^{r+1} = b_{ij}^r + \hat{u}_{j|i}v_j^r	  \\
		&c_{ij} = \frac{exp(b_{ij})}{\sum_k exp(b_{ik})}\\
		\end{aligned}
	\right.
	\label{caps2}
\end{equation}
The original dynamic routing uses the inner product to measure the consistency of neuron activity before and after the synapse. When extended to the SNNs, the information in the network is represented in a spiking vector. Compared with the real-valued artificial neuron, in addition to the part and the whole relationship between the low-level capsule and the high-level capsule in the spatial domain, there is also the spike timing information to describe the causal relationship between spike trains in the temporal domain. 
\begin{align}
\label{hebb}
c_{ij} = c_{ij} + \eta \sum \delta_{pre} \delta_{post}
\end{align}

Suppose only the original dynamic routing is extended to the SNNs, as shown in Eq.~\ref{hebb}. In that case, it equals to Hebb learning~\cite{hebb2005organization}, only considering the activity of neurons before and after synapses, ignoring the relative order of spikes firing between them, and missing the temporal information. $\eta$ is the learning rate. Moreover, since the SNN uses discrete spikes to transmit information, the sparse spike firing activity will make the weight of network not updated for a long time. Therefore, we consider extending the Hebb-like learning rule to the STDP learning rule, which we call STDP routing, to strengthen the connections between the low-level and high-level capsules in spatial and temporal domains. 

In order to simplify the calculation, we use synaptic traces $x_{pre}$ to record the spike information of neurons. The arrival of each pre-synaptic spike leaves a trace, which is updated to 1 when there is a spike and decays exponentially when there are no spikes. The details can be seen in Eq.~\ref{trace} where $\tau$ denotes the time constant.
\begin{align}
\label{trace}
\tau \frac{x_{pre}}{dt} = -x_{pre}
\end{align}

The spike traces encode the temporal relationship of neuronal spikes before and after the synapse in a positive time window. As shown in Fig.~\ref{routing}, if the pre-synaptic spike triggers the post-synaptic neuron within the time window, the connection will be strengthened. Otherwise, it will be weakened. When the pre-synaptic spikes arrive, the intensity of the change depends on the time difference between the pre-synaptic spike and the post-synaptic spike. We use the one-sided STDP shown in the Eq.~\ref{stdp}.
\begin{align}
	\label{stdp}
	c_{ij} = c_{ij} + \eta (x_{pre}-\chi_{offset}) \delta_{post}
\end{align}

For the parameters of other parts of the Spiking CapsNet, thanks to the proposal of the surrogate gradient, we can directly use the backpropagation algorithm to update their weights. We use the gradient of the rectangular function to replace that of the original threshold function~\cite{wu2018spatio}, which is shown in Eq.~\ref{gra}.

\begin{align}
	\label{gra}
	\frac{\partial \delta(V)}{\partial V} = \begin{cases}
	1 \quad if -\frac{1}{2}\leq V-V_{th} \leq \frac{1}{2}\\ 0 \quad else
	\end{cases}
\end{align}
\subsection{The whole Spiking CapsNet model}
\label{architecture}

\begin{figure*}[!htbp]
	\centering
	\includegraphics[scale=0.55]{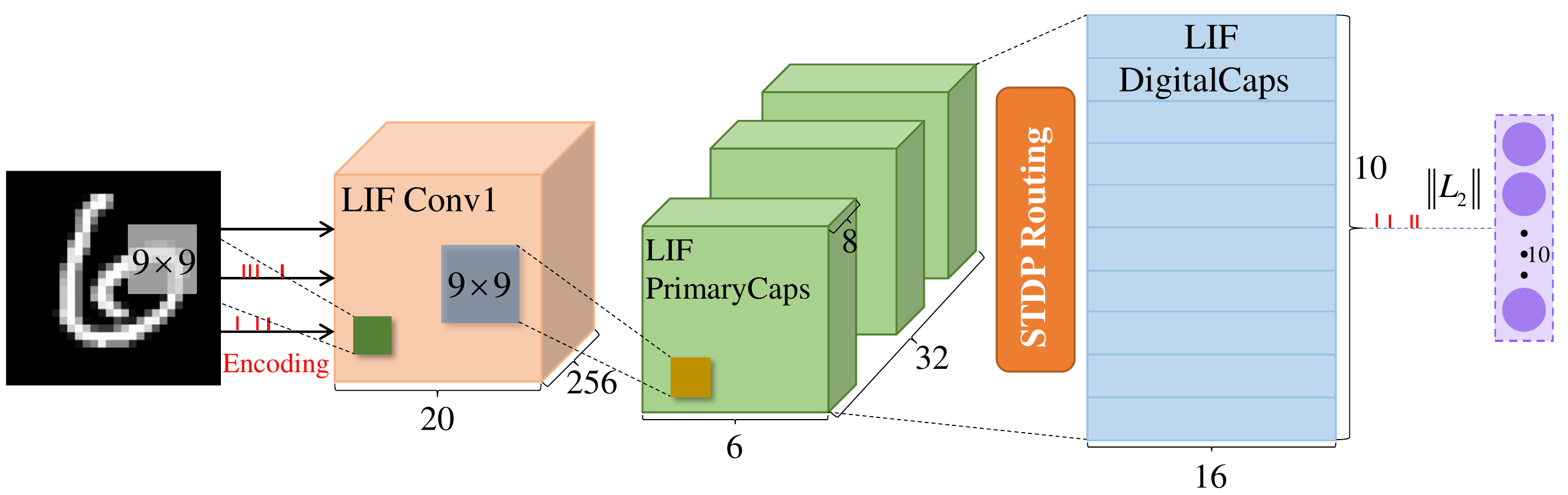}
	\caption{The architecture of Spiking CapsNet}
	\label{SpikingCaps}
\end{figure*}
The whole pipeline of our Spiking CapsNet is shown in Fig.~\ref{SpikingCaps}, where we combine the characteristics of SNNs and CapsNet efficiently. 

We use the $9 \times 9$ kernel shape with 256 output channels for the first convolutional layer. The PrimaryCaps is a convolutional capsule layer with 32 capsules. Each capsule contains eight spiking units. The PrimaryCaps further extracts features, and after transformation, 6*6*32 spiking capsules of length eight are obtained. The final DigitalCaps has a 16-dimensional capsule for each digital class after calculation.

The squash function can shrink the short and long vectors to 0 and 1, respectively. However, in Spiking CapsNet, spikes are used to propagate information between capsules, only 0 and 1. The firing rate of each capsule unit is naturally between 0-1. As a result, the squash function can be withdrawn from our model so as the softmax operation. 

For the output form of DigitalCaps, we have designed two different modes, called Spiking CapsNet Norm and Spiking CapsNet FC. For Spiking CapsNet Norm, we adopt the same operation as in the original CapsNet, using the L2 Norm of the membrane potential of each class as the final output. For Spiking CapsNet FC, we use a fully connected layer to take the output spikes of the DigitalCaps layer as input and predict the probability of each class directly. And we adopt the same mean squared error (MSE) loss function used in~\cite{wu2018spatio}. 

\section{Experiments}
In this section, we conduct experiments on several famous datasets to verify the performance of our model. Also, we test the robustness to the noise with different types of noises. Finally, we test its fitness to affine transformations. All the experiments are conducted on NVIDIA TITAN RTX with the Pytorch framework~\cite{paszke2017automatic}. The initial synaptic weights are set with the default method of Pytorch. The Adam optimizer ~\cite{kingma2014adam} and learning rate decay strategy is adopted, decreasing to 0.3*$lr$ at the epoch. The hyperparameters used in these experiments are listed in Tab.~\ref{para}. For the traditional classification performance and noise robust experiments, we initialize the parameters of the DigitalCaps layer with the following scheme:
\begin{align}
	\mathbf{W} \in U[-\sqrt{\frac{3}{mn}}, \sqrt{\frac{3}{mn}}], \quad \mathbf{b} \in U[-\sqrt{\frac{3}{mn}}, \sqrt{\frac{3}{mn}}]
\end{align}
where $m$ and $n$ are the length of the input and output capsules.

\begin{table}[h]
	\centering
	\begin{tabular}{lr}
		\toprule
		Parameter & Value\\
		\midrule
		Mini Batch Size & 50 \\
		Time Window for Inference and Training & 5ms\\
		Threshold of LIF Neuron & 0.5\\
		Decay Constant of LIF Neuron & 0.2\\
		Decay Constant of Synaptic Trace & 1.5\\
		Number of Epoches in Training & 50\\
		Learning Rate of STBP & 1e-3\\
		\bottomrule
	\end{tabular}
	\caption{Hyperparameters Used in Our Experiments}
	\label{para}
\end{table}
\subsection{Traditional Classification Performance}
\begin{figure*}[!htbp]
	\centering
	\subfigure[]{
		\label{matrixmnist}
		\includegraphics[scale=0.32]{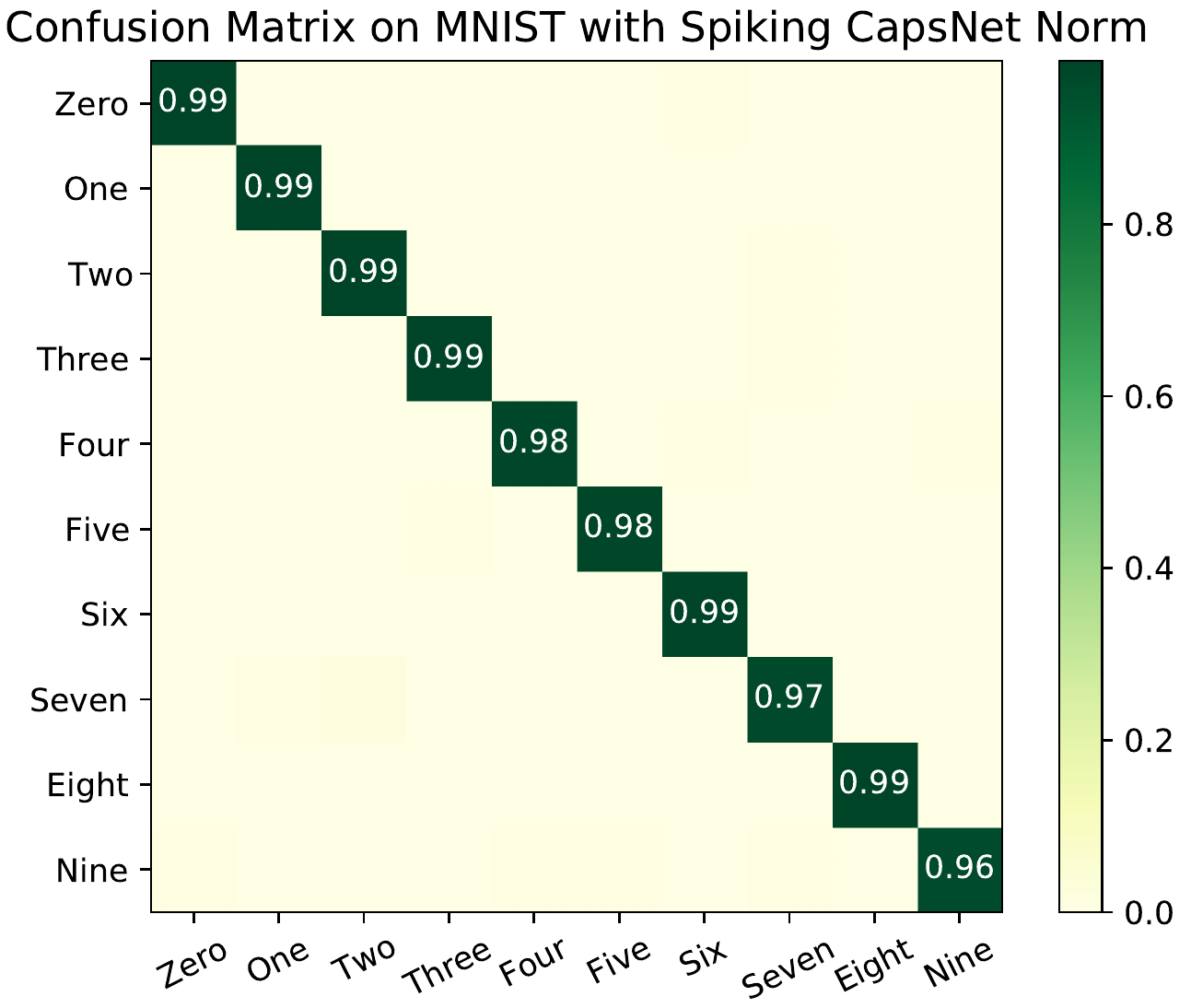}}
	\subfigure[]{
		\label{matrixmnistfc}
		\includegraphics[scale=0.32]{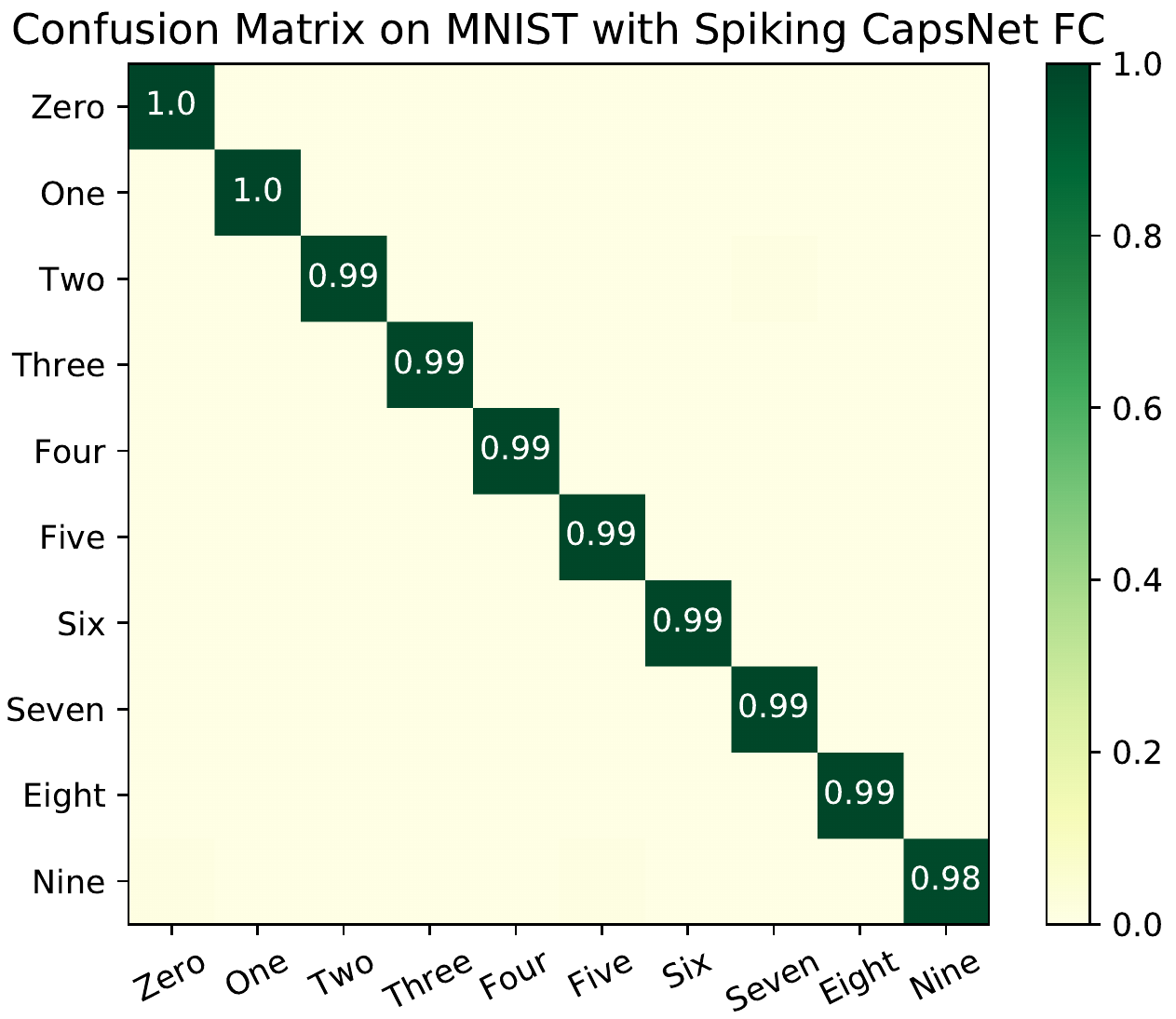}}
			\subfigure[]{
		\label{matrixFashion}
		\includegraphics[scale=0.31]{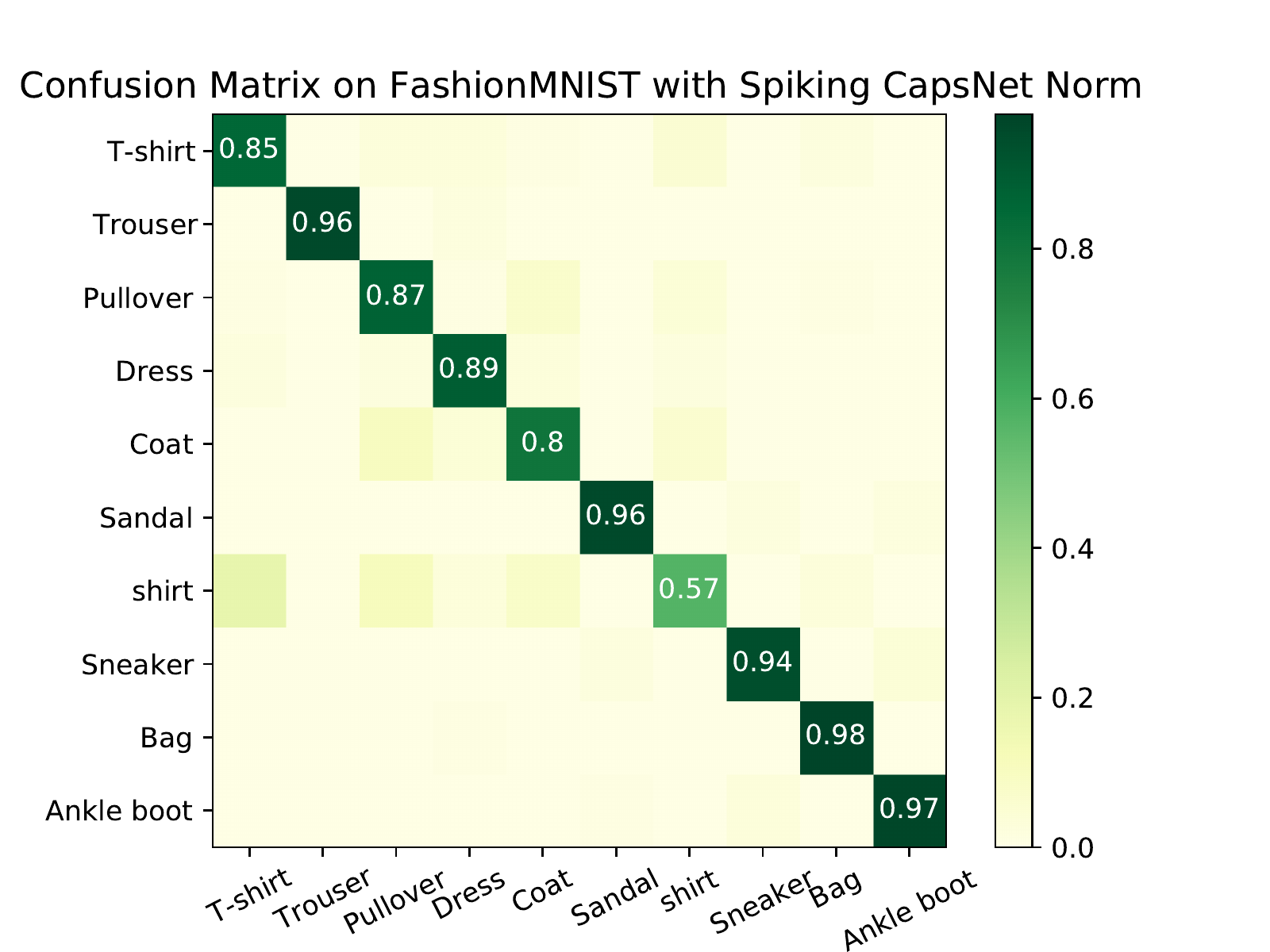}}
	\subfigure[]{
		\label{matrixFashionfc}
		\includegraphics[scale=0.31]{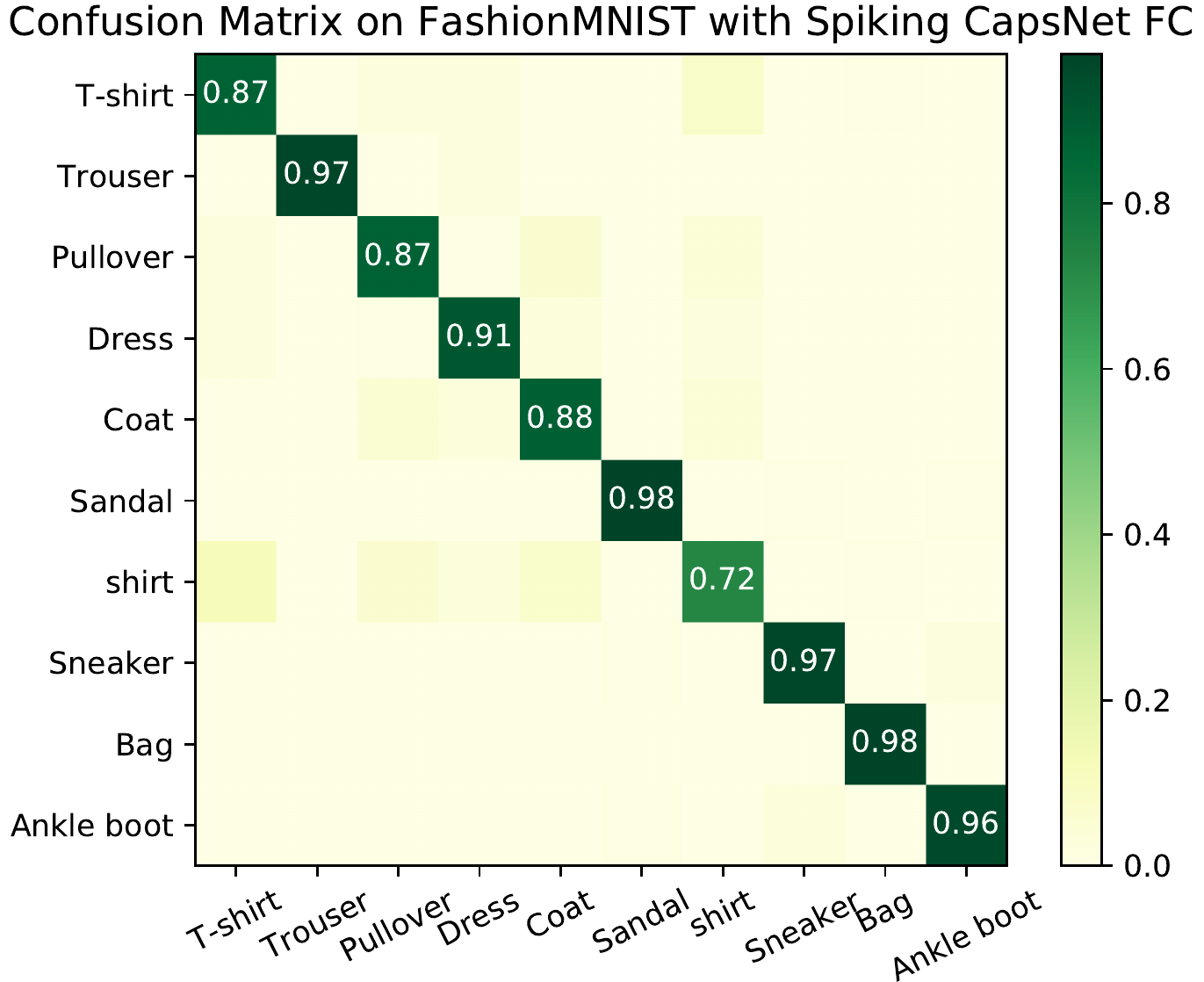}}
	
	\caption{The confusion matrix on MNIST and FashionMNIST with Spiking CapsNet Norm and Spiking CapsNet FC.}
	\label{ConMatrix}
\end{figure*}

We first verify our Spiking CapsNet on the MNIST dataset. The MNIST dataset contains 60,000 training samples and 10,000 test samples, describing grayscale images of handwritten digits with a size of 28*28 from 0-9. In order to improve the recognition accuracy, we normalize MNIST to make it obey the standard normal distribution. Moreover, we use the direct input encoding strategy. As described in Section Capsule Neural Networks, Spiking CapsNet has the same network structure as CapsNet. The classification accuracy of the Spiking CapsNet Norm and Spiking CaspNet FC is shown in Fig.~\ref{matrixmnist} and Fig.~\ref{matrixmnistfc}.
After that, we conduct further verification on the FashionMNIST dataset, a complex version of the MNIST dataset. FashionMNIST is composed of 10 kinds of clothing images, and its data size and format are the same as MNIST. The recognition accuracy rate of Spiking CapsNet is shown in Fig.~\ref{matrixFashion} and Fig.~\ref{matrixFashionfc}. Fig.~\ref{ConMatrix} demonstrate that our proposed Spiking CapsNet could correctly classify each category with little confusion. 
For the T-shirt and shirt in Fig.~\ref{matrixFashion} and Fig.~\ref{matrixFashionfc}, the Spiking CapsNet cannot classify them well. For humans, it is also a challenging task. Also, the Spiking CapsNet FC can better fit the data, combining the output capsule features to achieve better classification performance than the feature length.

\begin{table}[!t]
	\centering
	\begin{threeparttable}  
	\resizebox{\columnwidth}{!}{
	\begin{tabular}{llllr}
		\toprule
		Dataset & Method & Structure & Learning & Performance( \%)\\
		\midrule
		& sym-STDP \cite{hao2020biologically} & Fc & STDP & 96.73\\
		& EMSTDP \cite{shrestha2019approximating} & Fc & STDP & 97.30\\
		& VPSNN \cite{zhang2018plasticity} & Fc & Equilibrium learning+STDP & 98.52\\
		& Spiking CNN \cite{tavanaei2018training} & Spiking CNN & Backpropagation+STDP & 98.60\\
		& GLSNN \cite{zhao2020glsnn} & Fc & Global Feedback+STDP & 98.62 \\
		MNIST& Spiking ConvNet \cite{diehl2015fast} & Spiking CNN & ANN-SNN Conversion & 99.10\\
		& STDBP \cite{zhang2020rectified} & Spiking CNN & Backpropagation & 99.40\\
		& STBP \cite{wu2018spatio} & Spiking CNN & Backpropagation & 99.42\\
		& Converted SNN \cite{rueckauer2017conversion} & LeNet & ANN-SNN Conversion & 99.44 \\
		& \textbf{Spiking CapsNet Norm (Ours)} & \textbf{CapsNet} & \textbf{STBP+STDP Routing} & \textbf{98.25}\\
		& \textbf{Spiking CapsNet FC (Ours)} & \textbf{CapsNet} & \textbf{STBP+STDP Routing} & \textbf{99.17}\\
		
		\midrule
		& sym-STDP \cite{hao2020biologically} & Fc & STDP & 85.31\\
		& EMSTDP \cite{shrestha2019approximating} & Fc & STDP & 86.10\\
		& HM2BP \cite{jin2018hybrid} & Fc & Backpropagation & 88.99\\
		& GLSNN \cite{zhao2020glsnn} & Fc & Global Feedback + STDP & 89.05 \\
		& STDBP \cite{zhang2020rectified} & Spiking CNN & Backpropagation & 99.10\\
		FashionMNIST& ST-RSBP \cite{zhang2019spike} & Spiking RNN & Backpropagation & 90.13\\
		& STBP \cite{wu2018spatio}\tnote{1} & Spiking CNN & Backpropagation & 92.67\\
		& Converted SNN \cite{rueckauer2017conversion}\tnote{1} & LeNet & ANN-SNN Conversion & 92.67 \\
		&\textbf{Spiking CapsNet Norm (Ours)}& \textbf{CapsNet} & \textbf{STBP+STDP Routing} & \textbf{87.86}\\
		 &\textbf{Spiking CapsNet FC (Ours)} & \textbf{CapsNet} & \textbf{STBP+STDP Routing} & \textbf{91.07}\\
		\bottomrule
	\end{tabular}}
	\begin{tablenotes}   
		\footnotesize           
		\item[1] Our own experimental result.
	\end{tablenotes} 
	\end{threeparttable}
	\caption{The Performance Comparisons of the Proposed Spiking CapsNet with other state-of-the-art algorithms}
	\label{comparison}
\end{table}
To demonstrate the superiority of our model, we further compare the performance of our Spiking CapsNet with other state-of-the-art algorithms. Our proposed Spiking CapsNet can achieve performance comparable to the best algorithm. As shown in Tab.~\ref{comparison}, for two tasks, the best accuracy of SNN is the converted SNN, which could losslessly inherit the performance of ANN. Our Spiking CapsNet reaches the accuracy of 99.17\% and 91.06\% with spike signals on MNIST and FashionMNIST datasets, which surpasses most of the algorithms, whether it is STDP or BP trained method.

\subsection{Noise Robustness}
We test the noise robustness of the proposed Spiking CapsNet by adding two representative types of noise, the Salt-Pepper noise and the gaussian noise to the input image.
\begin{figure*}[!htbp]
	\centering
	\subfigure[Salt-Pepper noise]{
		\label{SPMNIST}
		\includegraphics[scale=0.3]{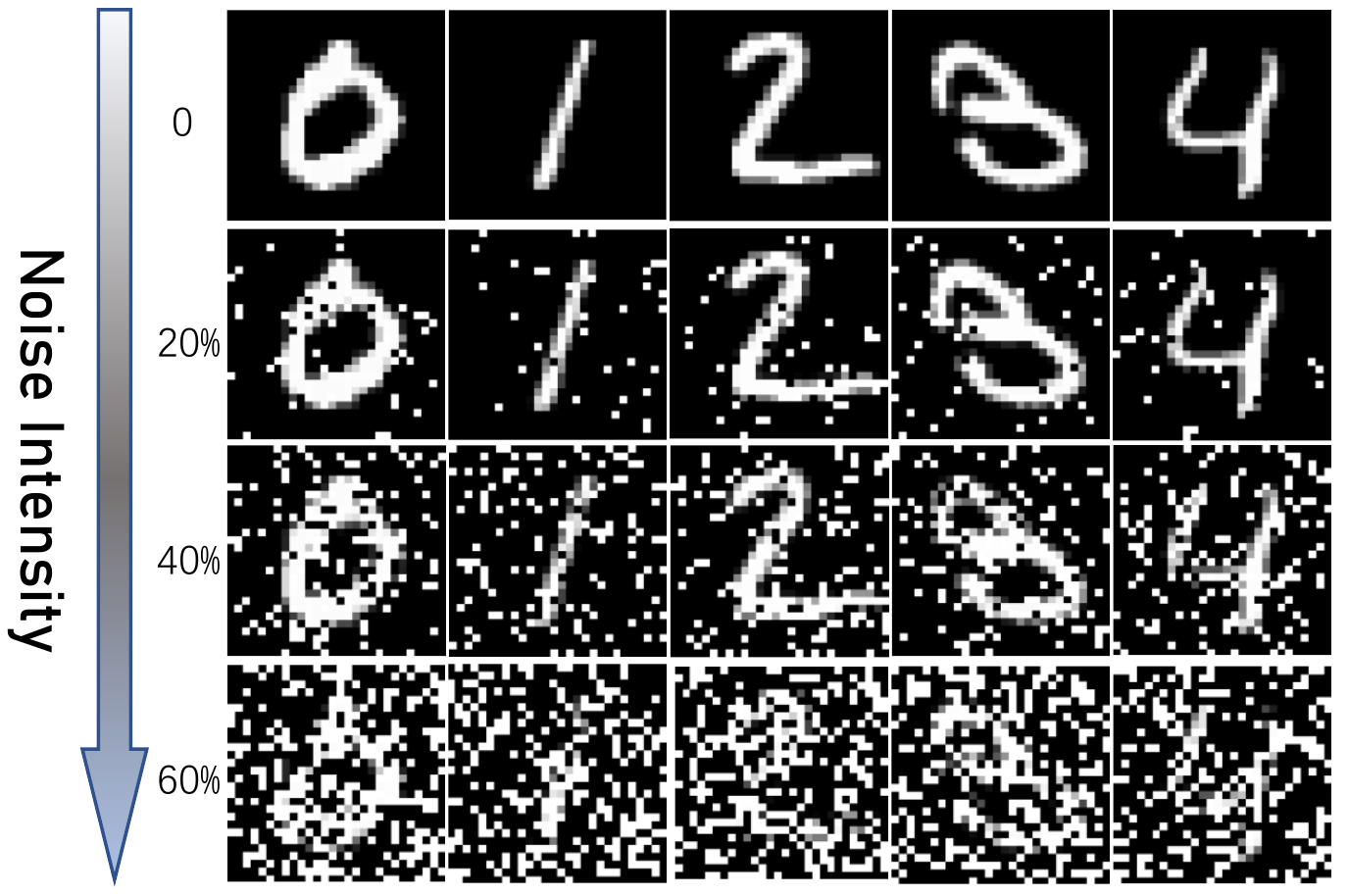}}
	\subfigure[Gaussian noise]{
	\label{GausMNIST}
	\includegraphics[scale=0.3]{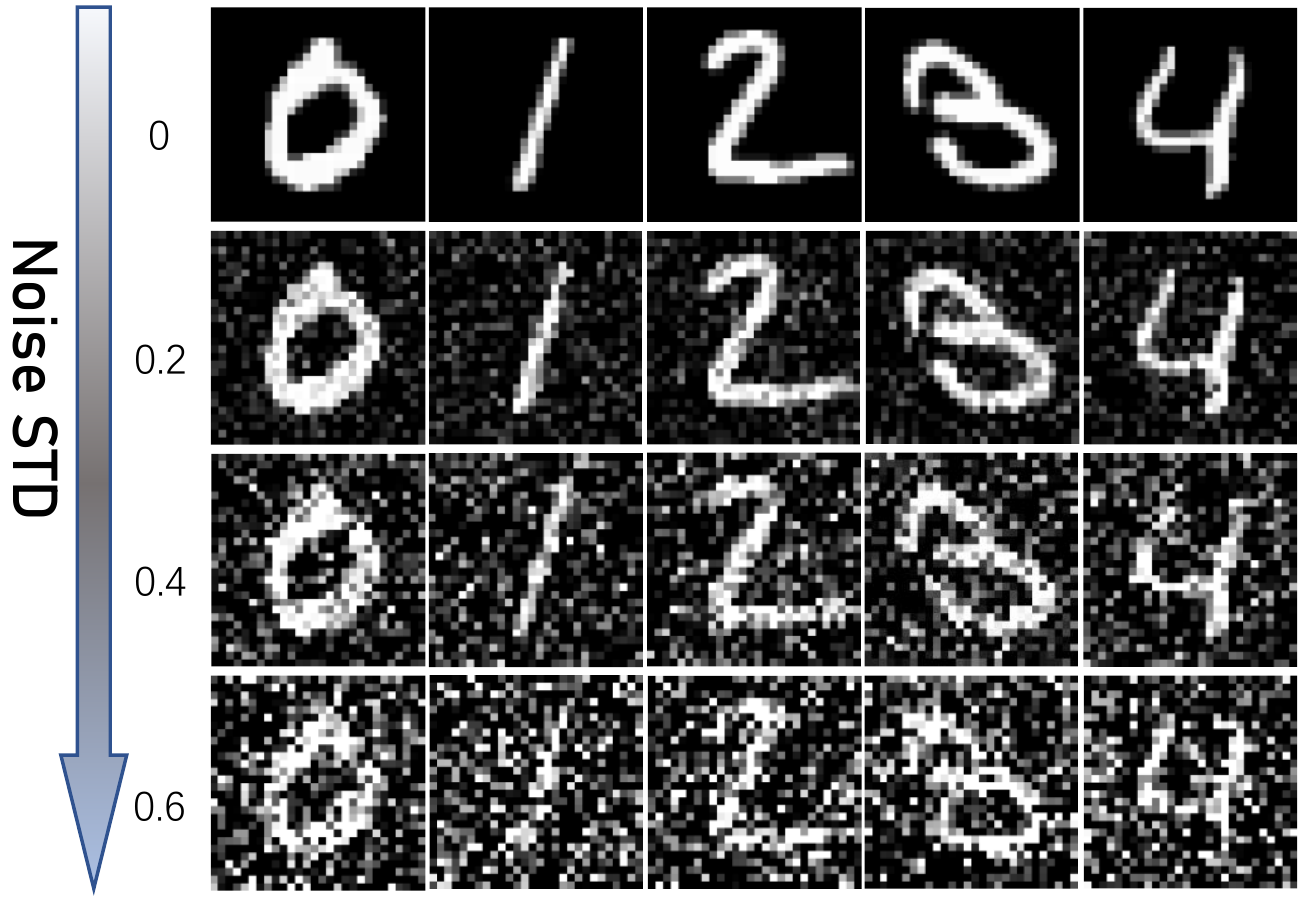}}
	\subfigure[Salt-Pepper noise]{
		\label{SPFashionMNIST}
		\includegraphics[scale=0.3]{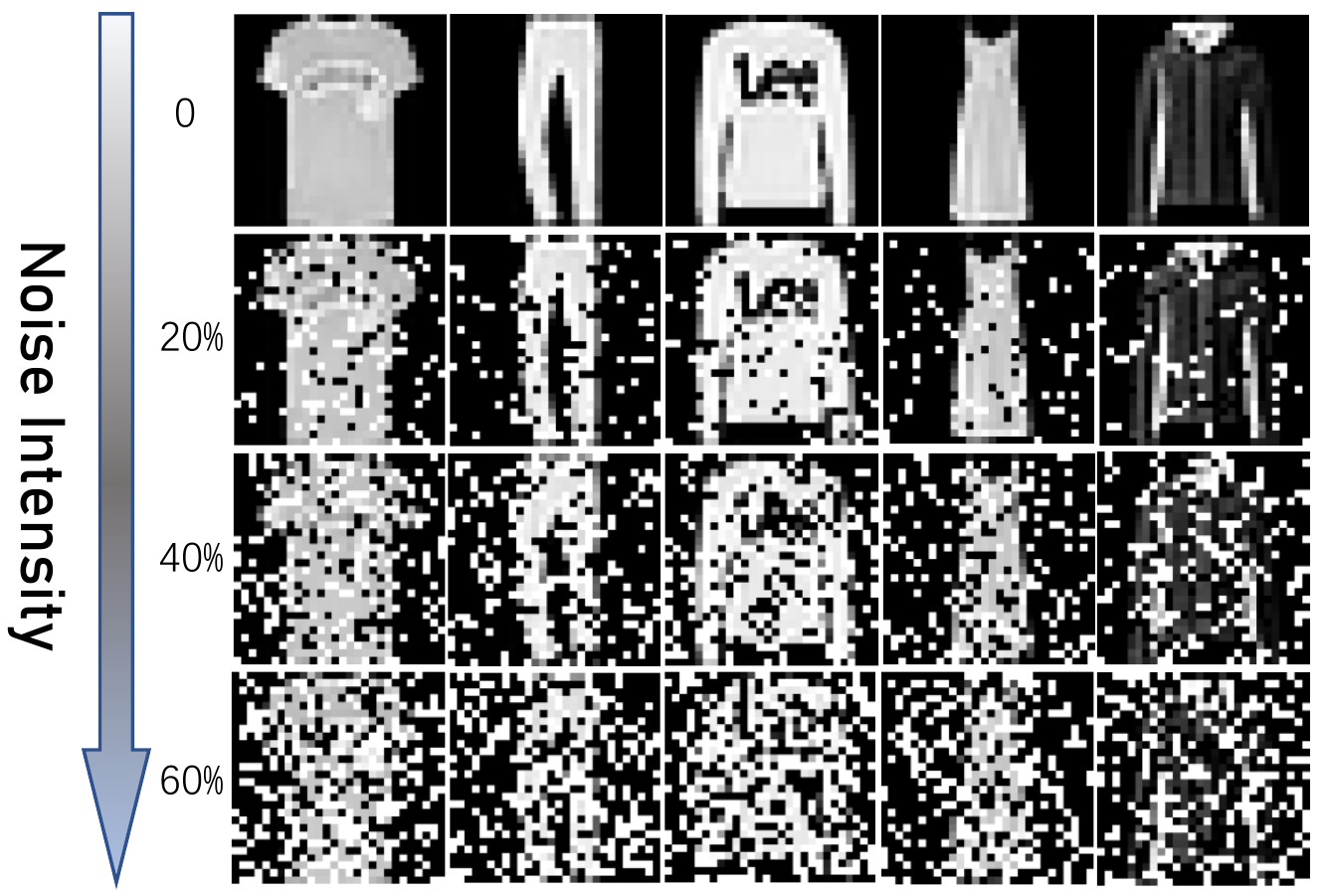}}
	\subfigure[Gaussian noise]{
		\label{GausFashionMNIST}
		\includegraphics[scale=0.3]{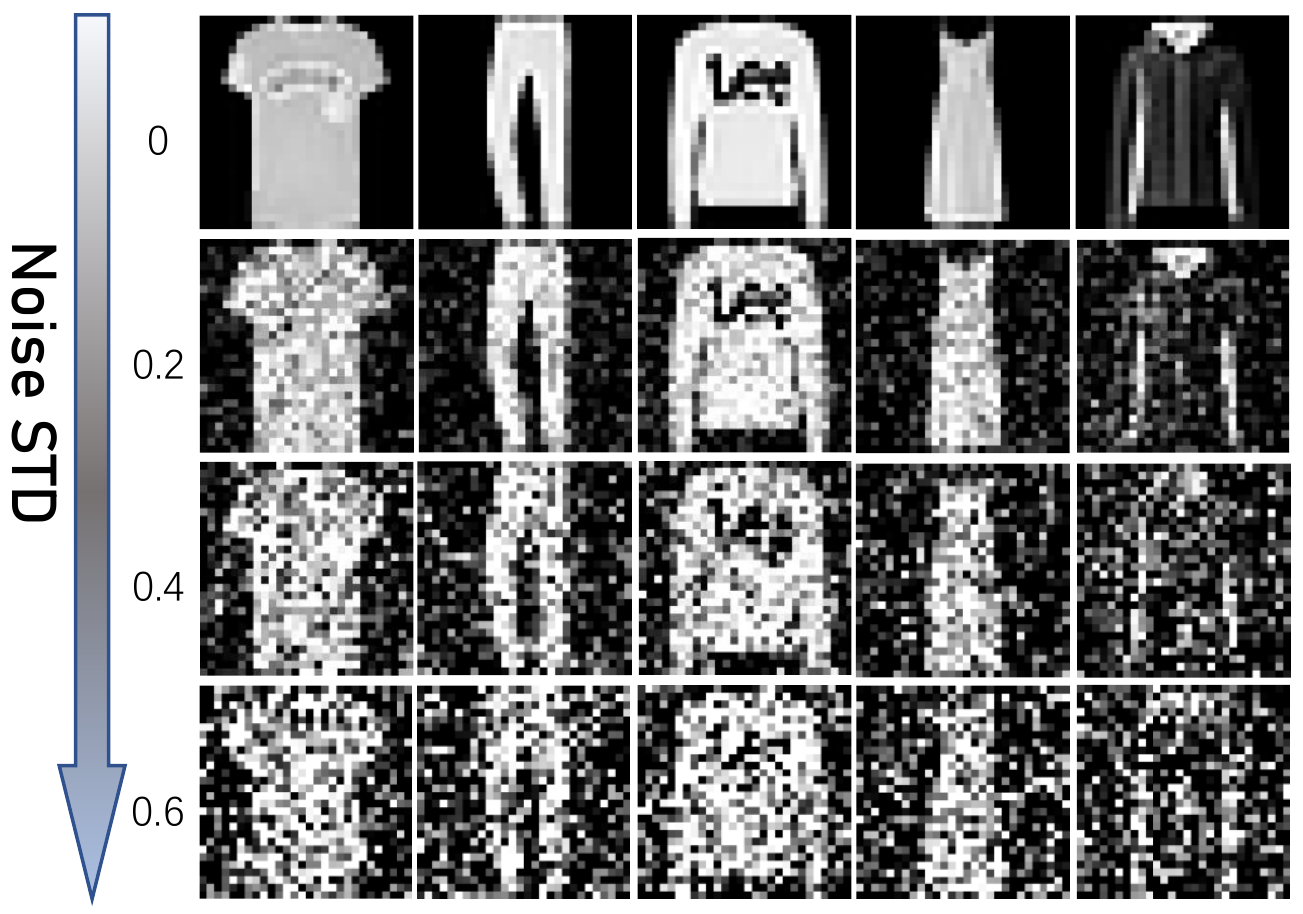}}
	\caption{MNIST and FashionMNIST Datasets with Salt-Pepper Noise and Gaussian Noise}
	\label{noise}
\end{figure*}

This paper tests different intensities of Salt-Pepper noise and different variance (mean value 0) Gaussian noise on MNIST and FashionMNIST datasets. The schematic diagram of the noised datasets is shown in Figure \ref{noise}. 

Spiking CapsNet FC uses a fully connected layer for output, focusing on the correspondence between features and labels. Spiking CapsNet Norm pays more attention to the internal representation of features and outputs using the L2 norm. When noises of different intensities and variances contaminate the data, the performance of the Spiking CapsNet Norm decreases more slowly. Detailed results are listed in Tab.~\ref{sprobustnetss} and Tab.~\ref{gaussrobustnetss}.

\begin{table}[!htbp]
	\centering
		\resizebox{\columnwidth}{!}{
			\begin{tabular}{llrrrrrrrrrr}
				\toprule
				Dataset & Model & 0 & 10\% & 20\% & 30\%  & 40\% & 50\% & 60\% & 70\%\\
				\midrule
				 MNIST & Spiking CapsNet Norm     & 98.25 & 96.36 & 89.54 & 77.44 & 63.29 & 51.14 & 39.16 & 28.90\\
				 MNIST & Spiking CapsNet FC & 99.17 & 96.74 & 86.77 & 69.87 & 52.91 & 44.64 & 32.98 & 20.71\\
				\midrule
				 FashionMNIST & Spiking CapsNet Norm     & 87.86 & 83.78 & 77.55 & 68.97 & 61.30 & 52.42 & 43.28 & 32.71\\
				 FashionMNIST & Spiking CapsNet FC & 91.06 & 82.35 & 67.98 & 50.38 & 36.90 & 26.63 & 20.91 & 16.34\\
				\bottomrule
		\end{tabular}}
	\caption{The Noise Robustness of Spiking CapsNet on MNIST and FashionMNIST Datasets with Salt-Pepper Noise. The numbers in the first line mean the intensity of noise.}
	\label{sprobustnetss}
\end{table}

\begin{table}[!htbp]
	\centering
	\resizebox{\columnwidth}{!}{
		\begin{tabular}{llrrrrrrrrrr}
			\toprule
			Dataset & Model & 0.0 & 0.1 & 0.2 & 0.3  & 0.4 & 0.5 & 0.6 & 0.7 \\
			\midrule
			MNIST & Spiking CapsNet Norm & 98.25 & 97.80 & 95.44 & 88.42 & 77.35 & 66.98 & 58.75 & 51.69\\
			MNIST & Spiking CapsNet FC & 99.17 & 98.80 & 92.74 & 77.03 & 63.32 & 51.69 & 44.54 & 38.88\\
			\midrule
			FashionMNIST & Spiking CapsNet Norm & 87.86 & 87.24 & 84.42 & 79.42 & 71.24 & 64.33 & 57.30 & 51.27\\
			FashionMNIST & Spiking CapsNet FC & 91.06 &  89.47 & 83.13 & 70.07 & 52.55 & 38.96 & 30.06 & 24.11\\
			\bottomrule
	\end{tabular}}
	\caption{The Noise Robustness of Spiking CapsNet on MNIST and FashionMNIST Datasets with Gaussian Noise. The numbers in the first line mean the STD of Gaussian noise, whose mean is zero.}
	\label{gaussrobustnetss}
\end{table}

Then we compare the noise robustness of Spiking CapsNet and other typical methods, including CapsNet, traditional DNN LeNet, BP-based SNN STBP, and conversion-based SNN. 
\begin{figure}[!htbp]
	\centering
	\subfigure[The performance on MNIST with Salt-Pepper noise.]{
		\label{SPMNISTacc}
		\includegraphics[scale=0.33]{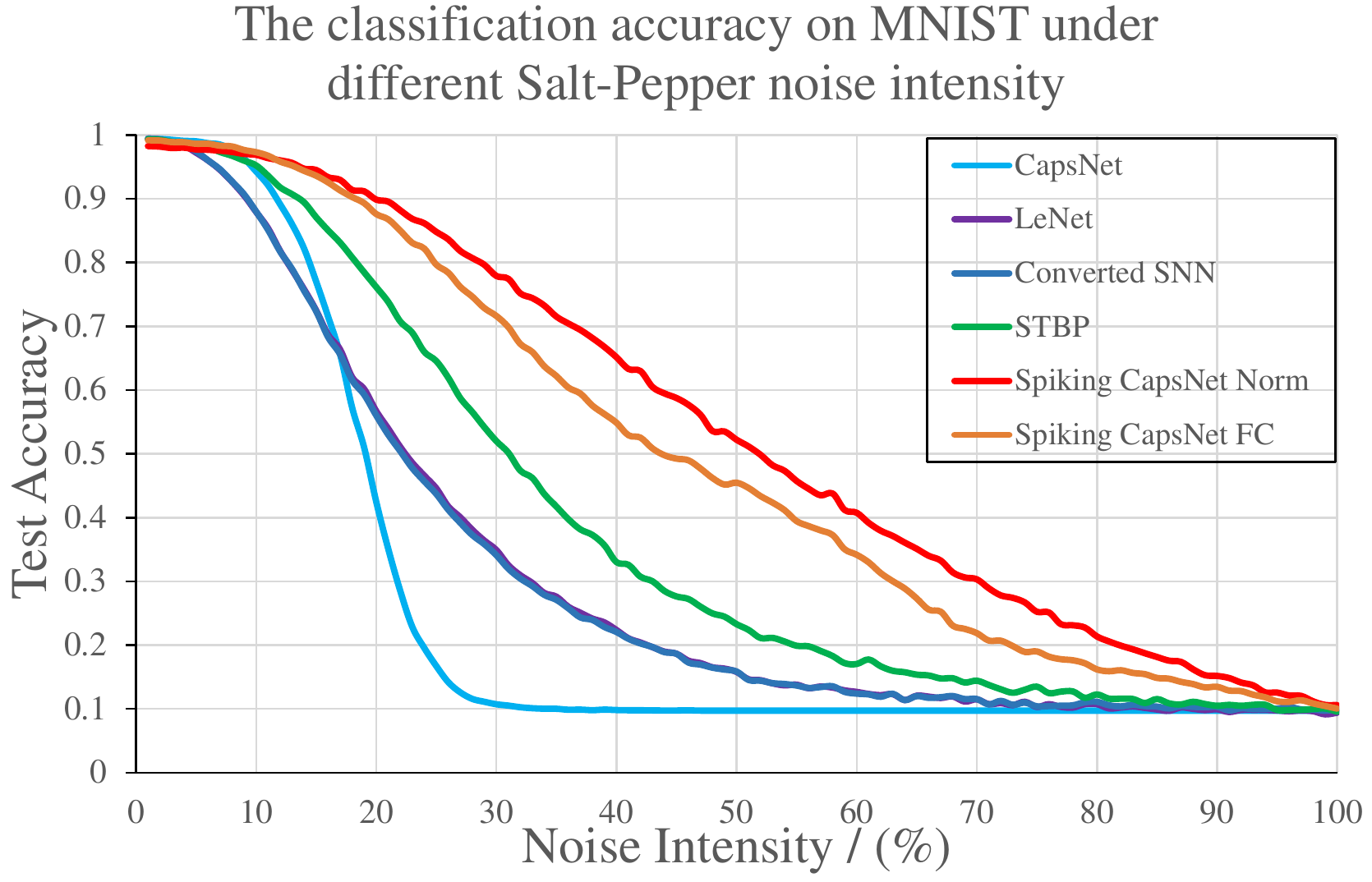}}
	\subfigure[The performance on FashionMNIST with Salt-Pepper noise.]{
		\label{SPFashionMNISTacc}
		\includegraphics[scale=0.33]{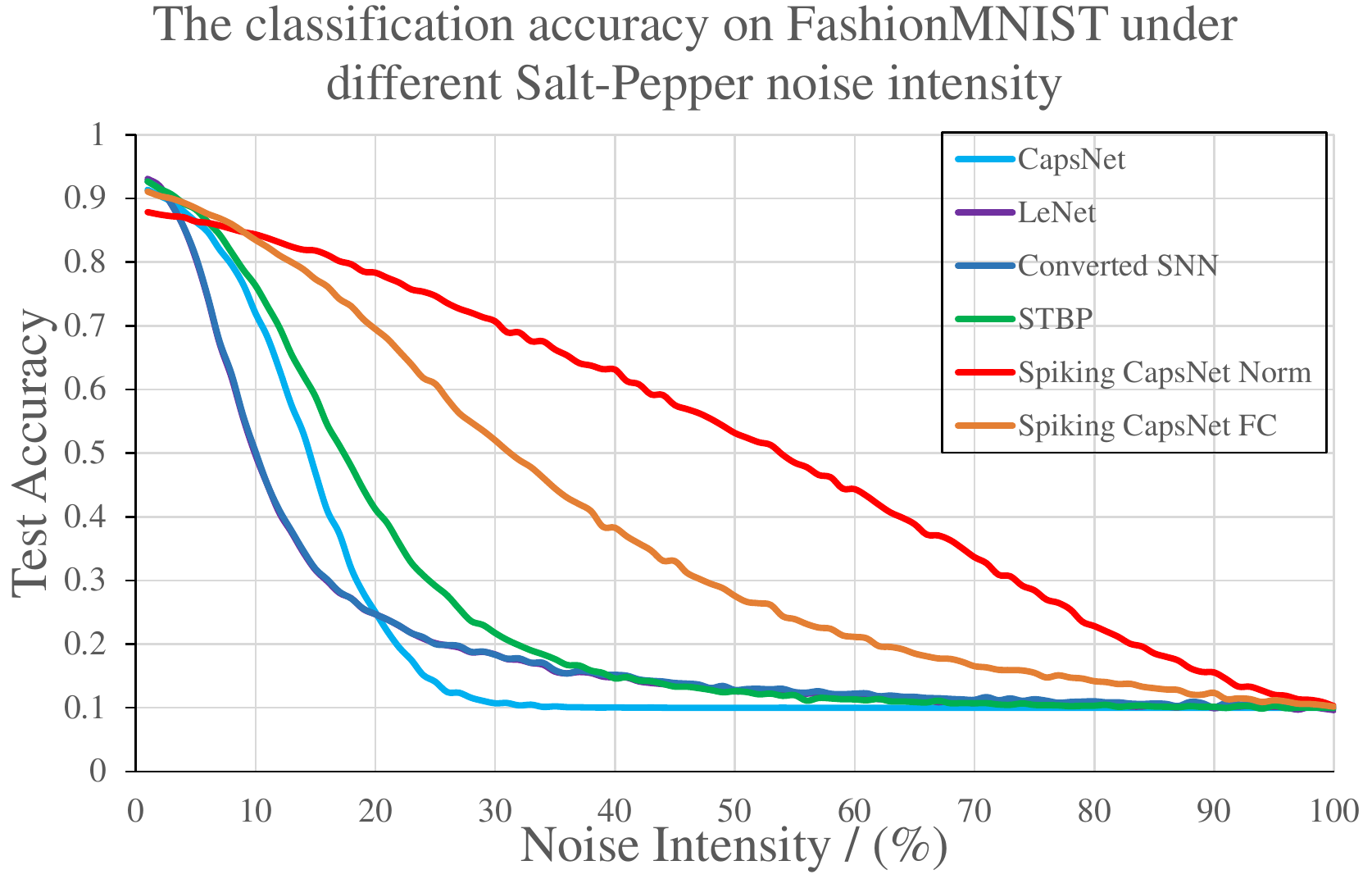}}
	\caption{The classification on MNIST and FashionMNIST Datasets with different Salt-Pepper Noise intensity.}
	\label{sprobustness}
\end{figure}
As shown in Fig.~\ref{sprobustness}, when contaminated by Salt-Pepper noise, for the MNIST dataset, the performance of other algorithms drops rapidly in the noise range of 20\% to 40\%. It maintains a low performance until the classification ability is wholly lost. For the FashionMNIST dataset, these algorithms hardly work properly after 30\% noise intensity. However, our Spiking CapsNet algorithm shows strong noise robustness when facing Salt-Pepper noise on the two data sets and can still maintain considerable accuracy before 40-70\% intensity noise. 

\begin{figure}[!htbp]
	\centering	
	\subfigure[The performance on MNIST with Gaussian noise.]{
		\label{GausMNISTacc}
		\includegraphics[scale=0.33]{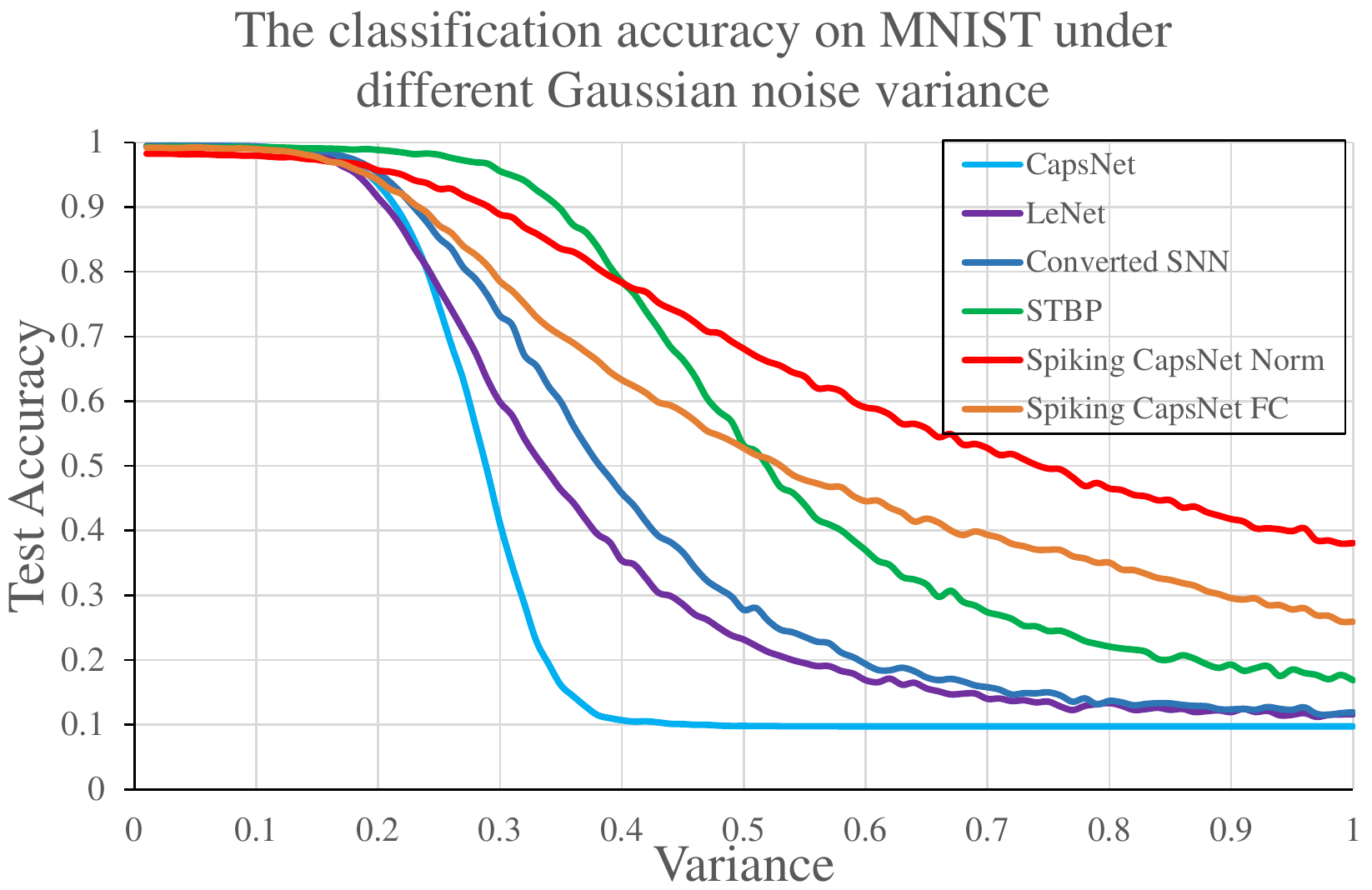}}
	\subfigure[The performance on FashionMNIST with Gaussian noise.]{
		\label{GausFashionMNISTacc}
		\includegraphics[scale=0.33]{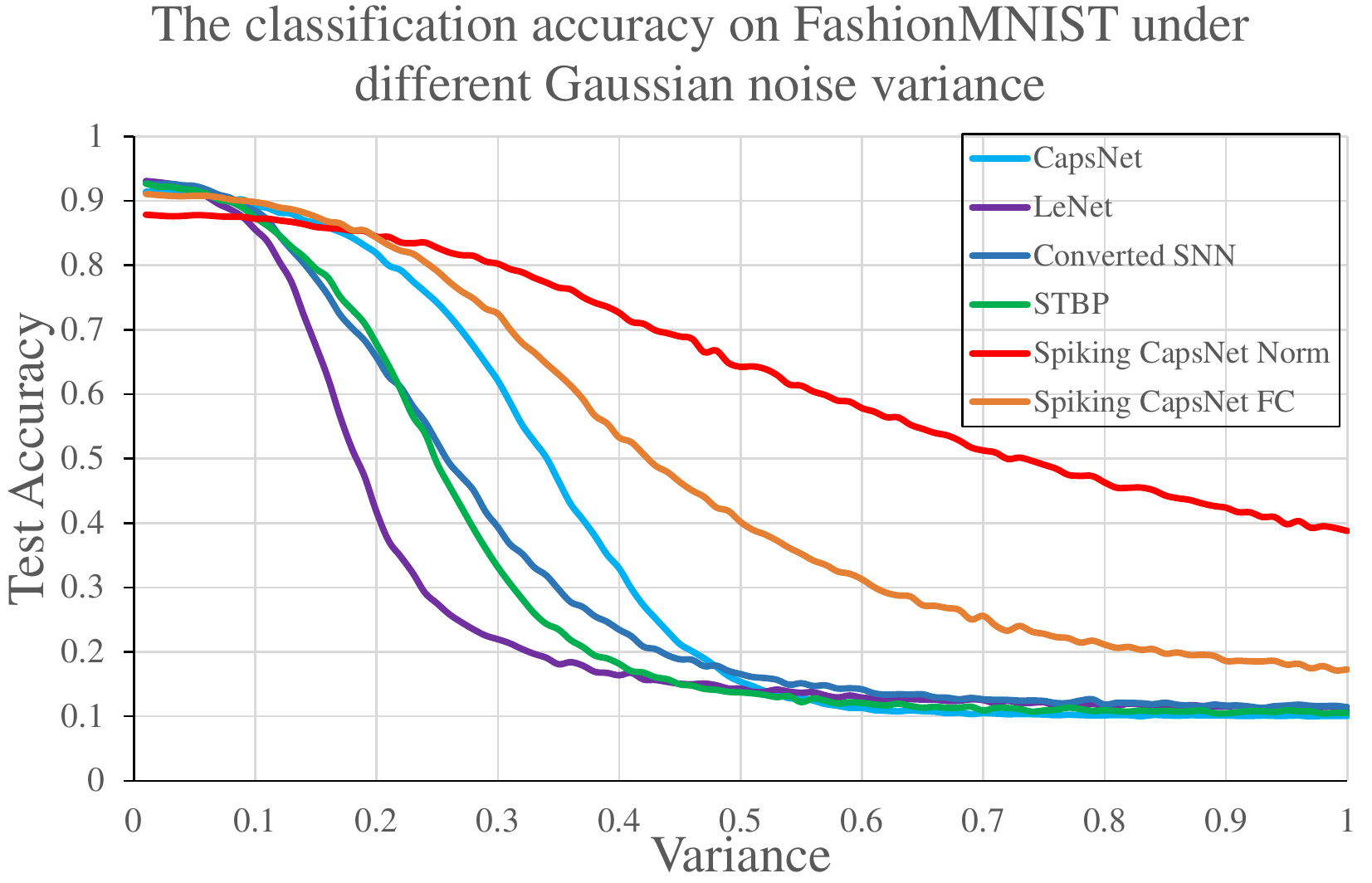}}
	\caption{The classification on MNIST and FashionMNIST datasets with different Gaussian Noise variance.}
	\label{gaussrobustness}
\end{figure}

Then, we use Gaussian noise with a mean value of 0 and a variance from 0 to 1. As can be seen from Fig.~\ref{gaussrobustness}, for FashionMNIST, our Spiking CapsNet can always maintain the accuracy of more than 40\%, while other algorithms lose their distinguishing ability after the noise variance is 0.5. Although on the MNIST dataset, the STBP algorithm performs well with low variance noise, overall, the robustness of our Spiking CapsNet is still the strongest. Note that when the Gaussian noise variance is 1, the input image does not entirely lose adequate information like 100\% intensity Salt-Pepper noise.

\subsection{Robustness to Affine Transformations}
\begin{figure}[!htbp]
	\centering
	\includegraphics[scale=0.45]{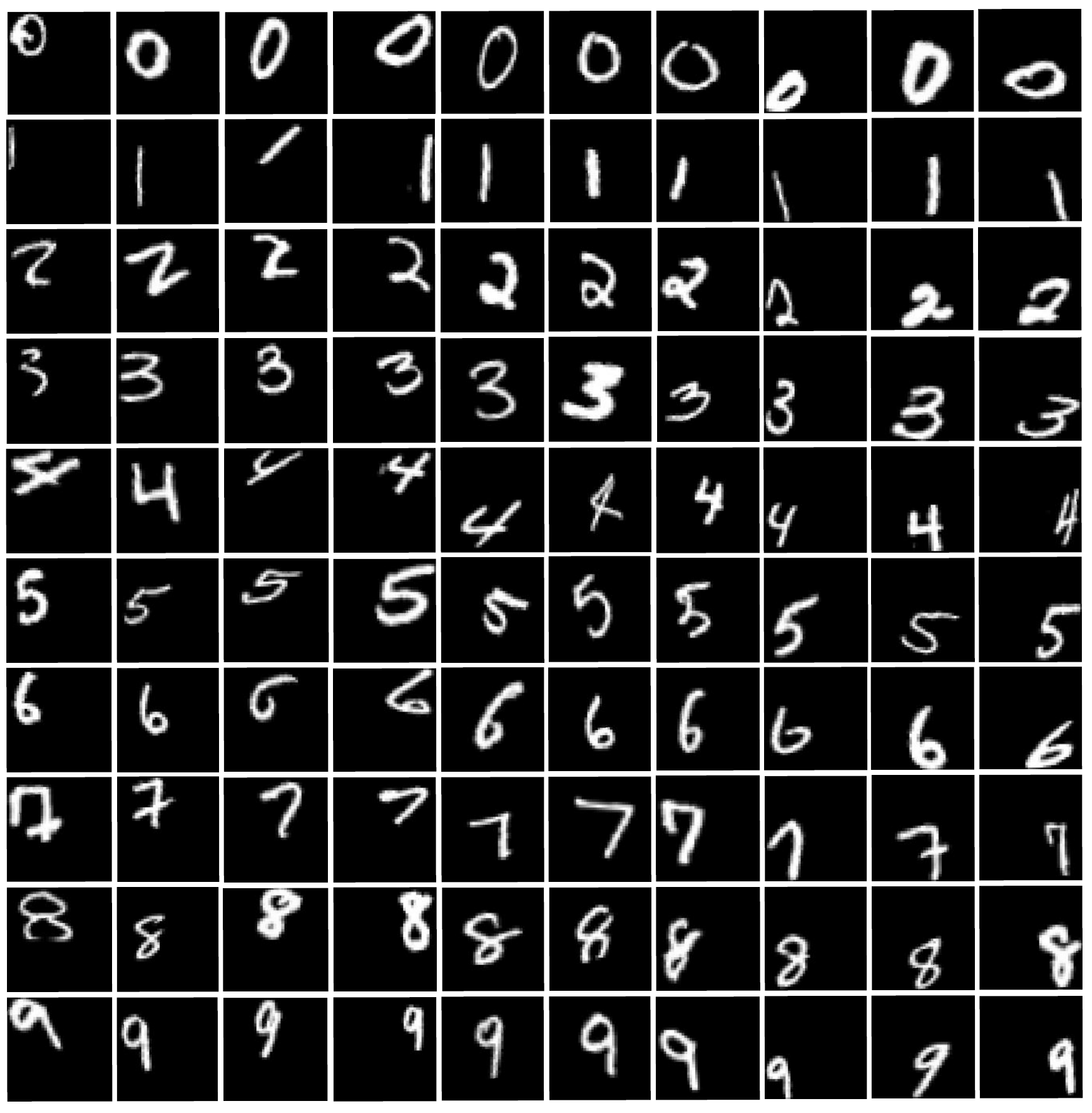}
	\caption{Some examples in AffNIST dataset.}
	\label{aff}
\end{figure}

To illustrate the generalization and invariance of our model for affine transformations, we replicate the experiment of~\cite{sabour2017dynamic}, we train the Spiking CapsNet FC on the MNIST training dataset and test it on the AffNIST test dataset. AffNIST is an expanded MNIST dataset, in which each sample is affine transformed on the corresponding sample in the MNIST dataset. The sample size is $40 \times 40$, and the details are shown in Fig~\ref{aff}.

In order to be consistent with AffNIST, we place the sample of the MNIST training dataset randomly on a background of $40\times 40$ pixels. We compared our model with a spiking convolutional neural network, a traditional convolutional neural network, and capsnet. In order to compare the robustness to affine transformation more fairly, we followed the same early stop operation in ~\cite{sabour2017dynamic}. We stopped training the model when the training accuracy achieved 97\%, 98\%, and 99\%, and the test accuracy on the AffNIST dataset respectively.

\begin{figure}
	\centering
	\includegraphics[scale=0.65]{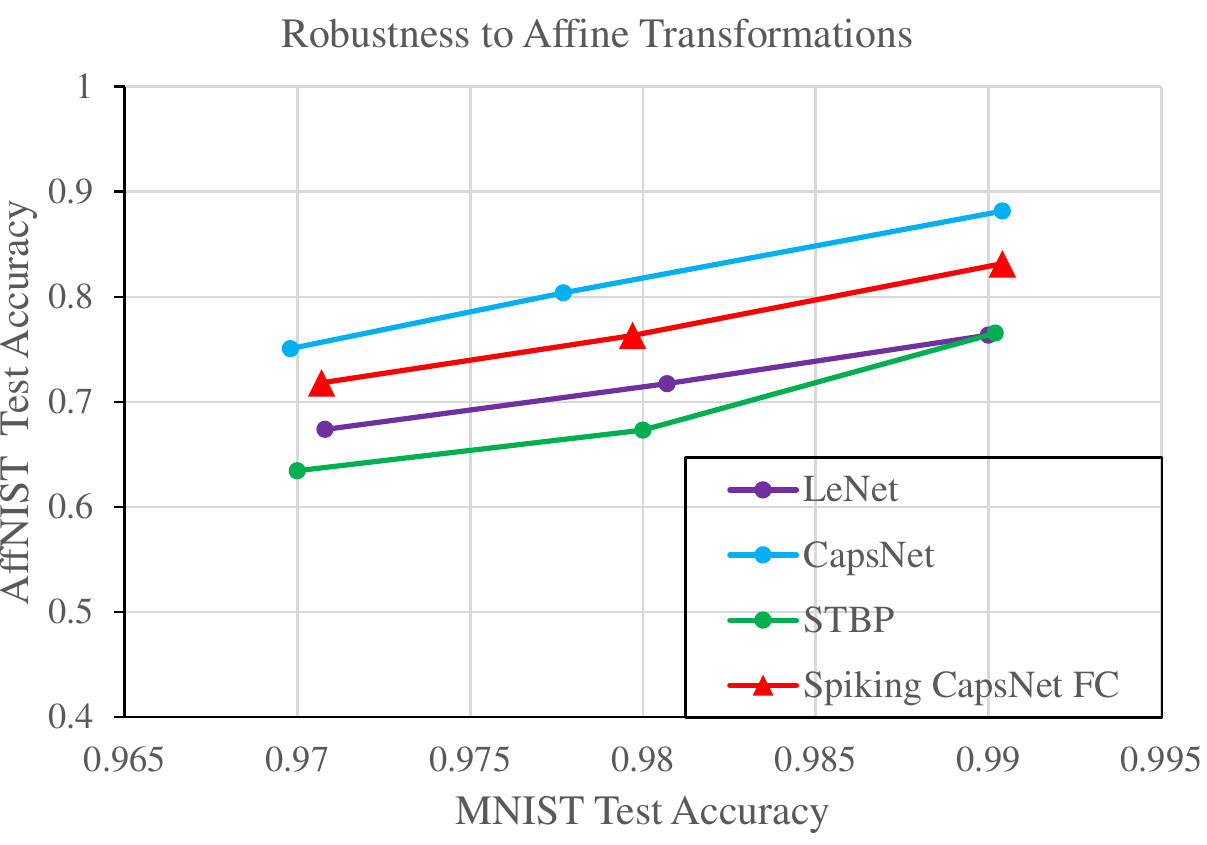}
	\caption{Convolutional Neural Networks vs Capsule Neural Networks.}
	\label{aff2}
\end{figure}

As shown in Fig.~\ref{aff2}, our Spiking CapsNet FC reaches 71.8\%, 76.3\%, and 83.2\%, while the spiking convolutional neural network and convolutional neural network only reaches or below 67.4\%, 71.1\%, and 76.4\%. Although compared with CapsNet, there is still a particular gap. We think it mainly comes from the fact that SNN is more difficult to train than DNN. This gap will be further reduced when a better SNN training algorithm is used in future work.

\section{Conclusion}
In this paper, the Spiking CapsNet is proposed by introducing the capsule concept into modelling the spiking neural network, which fully combines their spatio-temporal processing capabilities. Meanwhile, we propose a more biologically plausible STDP routing algorithm that fully considers the part and the whole spatial relationship between the low-level capsule and the high-level capsules, as well as the spike firing time order between the post-synaptic and post-synaptic neurons. The coupling ability between low-level and high-level spiking capsules is further improved. Our model shows strong performance and robustness on multiple datasets. We have tested on the MNIST and FashionMNIST datasets. Compared with other state-of-the-art SNN algorithms, our Spiking CapsNet shows comparable performance. In addition, it shows strong robustness to different Salt-Pepper noises and Gaussian noises. Finally, our model has strong adaptability to the spatial affine transformation.

\section*{Acknowledgement}
This work is supported by the National Key Research and Development Program (Grant No. 2020AAA0104305), the Strategic Priority Research Program of the Chinese Academy of Sciences (Grant No. XDB32070100), the Beijing Municipal Commission of Science and Technology (Grant No. Z181100001518006), the Key Research Program of Frontier Sciences, Chinese Academy of Sciences (Grant No. ZDBS-LY-JSC013), and the Beijing Academy of Artificial Intelligence.
\bibliography{aaai22}

\end{document}